# A Fuzzy Clustering Model for Fuzzy Data with Outliers

M.H.Fazel Zarandi, Zahra.S.Razaee
Department of Industrial Engineering,
Amirkabir University of Technology, Tehran, Iran
zarandi@aut.ac.ir, razaei@aut.ac.ir

May 2010

**Abstract**

In this paper a fuzzy clustering model for fuzzy data with outliers is proposed. The model is based on Wasserstein distance between interval valued data which is generalized to fuzzy data. In addition, Keller's approach is used to identify outliers and reduce their influences. We have also defined a transformation to change our distance to the Euclidean distance. With the help of this approach, the problem of fuzzy clustering of fuzzy data is reduced to fuzzy clustering of crisp data. In order to show the performance of the proposed clustering algorithm, two simulation experiments are discussed.

## 1 Introduction

Clustering is a division of a given set of objects into subgroups or clusters, so that objects in the same cluster are as similar as possible, and objects in different clusters are as dissimilar as possible. From a machine learning perspective, clustering is an unsupervised learning of a hidden data concept (Berkhin [3]). In conventional (hard) clustering analysis, each datum belongs to exactly one cluster, whereas in fuzzy clustering, data points can belong to more than one cluster, and associated with each datum is a set of membership degrees.

Fuzzy data are imprecise data obtained from measurements, human judgements or linguistic assessments. In cluster analysis, when there is simultaneous uncertainty in both the partition and data, a fuzzy clustering model for fuzzy data should be applied (D'Urso and Giordani [11]).

In recent literature, there are several works regarding the fuzzy clustering of fuzzy data. Hathaway et al. [18] and Pedrycz et al. [29] introduced models that convert parametric or non-parametric linguistic variables into generalized coordinates before performing fuzzy c-means clustering. Yang and Ko [35] presented a fuzzy k-numbers clustering model that uses a squared distance between each pair of fuzzy numbers. Yang and Liu [38] extended the Yang and Ko work and proposed a fuzzy k-means clustering model for conical fuzzy vectors. Yang et al. [36] proposed a fuzzy K-means clustering model for handling both symbolic and fuzzy data. Hung and Yang [20] proposed an alternative fuzzy k-numbers clustering model which is based on exponential-type distance measure. D'Urso and Giordani [11] proposed a weighted fuzzy c-means clustering model which considers fuzzy data with a symmetric LR membership function.

In this paper, we first propose a new distance measure for comparison of fuzzy data. On account of the fact that all the $\alpha$-cuts of fuzzy data are intervals, we obtain the distance between two fuzzy data from the distances between their $\alpha$-cuts. To this purpose, a special case of Wasserstein distance is utilized. The choice of $\alpha$-cuts is motivated by the fact that, fuzzy data with different shapes can be used. After introducing our distance, we use it for fuzzy clustering of fuzzy data. Moreover, with the help of Keller's [22] approach, an additional weighting factor is added for each datum to identify outliers and reduce their effects. In other approach, by definition of a transformation, triangular fuzzy data are changed to crisp data. With this novel approch, after applying the transformation, any fuzzy clustering model for crisp data can be used. Furthermore, for determining the optimal number of clusters, there is no need to define a cluster validity index for fuzzy data. The ones existing in literature for crisp data can be applied.

The rest of the paper is organized as follows. In Section 2, the concept of LR-type fuzzy data is introduced. Some related works regarding metrics for fuzzy data are reviewed in Section 3. We propose a distance measure for fuzzy data based on Wasserstein Metric in Section 4; by means of this distance and following Keller's approach, we propose a fuzzy clustering model for fuzzy data with outliers. Then, by defining of a new transformation, we change the fuzzy data to crisp data and for the sake of comparison, we again use Keller's algorithm (Section 5). Successively, in Section 6, the results of two simulation experiments are discussed. Finally, conclusions and future works are presented in Section 7.

## 2 LR-type fuzzy data

The LR-type fuzzy data represent a general class of fuzzy data. When we are dealing with univariate LR fuzzy data, this kind of data can be shown by a vector of LR-fuzzy numbers. In the more general case of multivariate analysis, we have a matrix of LR-fuzzy numbers (De Oliveria,Pedrycz [9]). To be more specific, let $L$ (and $R$) be a decreasing shape function, which map $\mathbb{R}^+ \to [0,1]$ with $L(0) = 1$; $L(x) < 1, \forall x > 0$; $L(x) > 0, \forall x < 1$; $L(1) = 0$ or ($L(x) > 0, \forall x$ and $L(+\infty) = 0$) (Zimmerman, [42]). Then, a fuzzy number $\widetilde{A}$ is of LR-type if for $c$, $l > 0, r > 0$ in $\mathbb{R}$,

$$\mu_{\widetilde{A}}(x) = \begin{cases} L(\frac{c-x}{l}) & \text{for } x \leq c, \\ R(\frac{x-c}{r}) & \text{for } x \geq c. \end{cases} \tag{1}$$

where, $c, l, r$ are the center, left and right spreads of $\widetilde{A}$, respectively. Symbolically we can write $\widetilde{A} = (c, l, r)_{LR}$.

In LR-type fuzzy numbers, the triangular fuzzy numbers (TFNs) are most commonly used. An LR-type fuzzy number $\widetilde{A}$ is called triangular fuzzy number if $L(x) = R(x) = 1 - x$, characterized by the following membership function:

$$\mu_{\widetilde{A}}(x) = \begin{cases} 1 - \frac{c-x}{l} & \text{for } x \leq c, \\ 1 - \frac{x-c}{r} & \text{for } x \geq c. \end{cases} \tag{2}$$

# 3 Related works

In the recent literature, there are some distance measures for fuzzy data. We review some of them in this section.

**Definition**: Considering two crisp sets $A, B \subseteq \mathbb{R}^k$, and a distance $d(x, y)$ where, $x \in A$ and $y \in B$, the Hausdorff distance is defined as follows:

$$d_H(A, B) = \max \left\{ \sup_{x \in A} \inf_{y \in B} d(x, y), \sup_{y \in B} \inf_{x \in A} d(x, y) \right\}. \tag{3}$$

According to the concept of $\alpha$-cuts, the Hausdorff metric $d_H$ can be generalized to fuzzy numbers $\widetilde{F}, \widetilde{G}$, where $\widetilde{F}$(or $\widetilde{G}$):$\mathbb{R} \to [0, 1]$:

$$d_\rho(\widetilde{F}, \widetilde{G}) = \begin{cases} \left[ \int_0^1 (d_H(F_\alpha, G_\alpha))^\rho \; d_\alpha \right]^{1/\rho} & \text{if } \rho \in [1, \infty) \\ \sup_{\alpha \in [0,1]} d_H(F_\alpha, G_\alpha) & \text{if } \rho = \infty, \end{cases} \tag{4}$$

where, the crisp set $F_\alpha \equiv \{x \in \mathbb{R}^k : F(x) \geq \alpha\}, \alpha \in [0, 1]$, is called the $\alpha$-cut of $\widetilde{F}$ (Näther,[26]).

Tran and Duckstein [33] proposed the following distance between two intervals:

$$\begin{aligned} d_{TD}(A, B) &= \int_{-\frac{1}{2}}^{\frac{1}{2}} \int_{-\frac{1}{2}}^{\frac{1}{2}} \left\{ \left[ \left( \frac{a+b}{2} \right) + x(b-a) \right] \right. \\ &\quad \left. - \left[ \left( \frac{u+v}{2} \right) + y(v-u) \right] \right\}^2 dx\, dy \\ &= \left[ \left( \frac{a+b}{2} \right) - \left( \frac{u+v}{2} \right) \right]^2 + \frac{1}{3} \left[ \left( \frac{b-a}{2} \right)^2 + \left( \frac{v-u}{2} \right)^2 \right]. \end{aligned} \tag{5}$$

Then, they used it to formulate their distance measure for fuzzy numbers, but $d_{TD}$ does not satisfy the reflexivity property (Irpino and Verde [21]):

$$\begin{aligned} d_{TD}(A, A) &= \left[ \left( \frac{a+b}{2} \right) - \left( \frac{a+b}{2} \right) \right]^2 + \frac{1}{3} \left[ \left( \frac{b-a}{2} \right)^2 + \left( \frac{b-a}{2} \right)^2 \right] \\ &= \frac{2}{3} \left( \frac{b-a}{2} \right)^2 \geq 0. \end{aligned} \tag{6}$$

A squared Euclidean distance between a pair of LR-type fuzzy data $\widetilde{A}_1 = (c_1, l_1, r_1)$ and $\widetilde{A}_2 = (c_2, l_2, r_2)$ , where $c$ denotes the center and $l, r$ indicate, respectively, the left and right spread, is defined by Yang and Ko [35]:

$$d^2_{YK}(\lambda, \rho) = (c_1 - c_2)^2 + [(c_1 - \lambda l_1) - (c_2 - \lambda l_2)]^2 + [(c_1 + \rho r_1) - (c_2 + \rho r_2)]^2, \quad (7)$$

where $\lambda = \int_0^1 L^{-1}(t)\,dt, \rho = \int_0^1 R^{-1}(t)\,dt$ are parameters that summarize the shape of the left and right tails of the membership function and $L, R$ are decreasing shape functions which were defined in Section 2.

# 4 The proposed distance for fuzzy data

In this section, we first present a new distance measure for interval-valued data, and then it is used to formulate the distance measure for fuzzy data. Let $I_i = [a_i, b_i]$, be an interval for $i = 1, 2$. We can parameterize $I_i$ as follows:

$$I_i(t) = a_i + t(b_i - a_i) \qquad 0 \le t \le 1. \quad (8)$$

If we represent $I_i$ by means of its midpoint $m_i = \frac{a_i+b_i}{2}$ and radius $\delta_i = \frac{b_i-a_i}{2}$, Eq.8 can be rewritten as follows:

$$I_i(t) = m_i + (2t - 1)\delta_i \qquad 0 \le t \le 1. \quad (9)$$

The distance measure between $I_1$ and $I_2$ can be defined as follows:

$$\begin{aligned} d^2(I_1, I_2) &= \int_0^1 [I_1(t) - I_2(t)]^2\,dt \\ &= \int_0^1 [(m_1 - m_2) + (\delta_1 - \delta_2)(2t - 1)]^2\,dt \\ &= (m_1 - m_2)^2 + \frac{1}{3}(\delta_1 - \delta_2)^2. \end{aligned} \quad (10)$$

This distance takes into account all the points in both intervals. Irpino and Verde [21] has derived Eq.10 from another point of view, using the Wasserstein distance. To be more specific, let $F_1$ and $F_2$ be distribution functions, the Wasserstein $L_2$ metric is defined as follows (Gibbs and Su [14]):

$$d_{Wass}(F_1, F_2) = \left\{ \int_0^1 \left(F_1^{-1}(t) - F_2^{-1}(t)\right)^2 dt \right\}^{1/2}, \quad (11)$$

where $F_1^{-1}$ and $F_2^{-1}$ are the quantile functions of the two distributions. If we assume $F_i$ for $i = 1, 2$ to be the uniform distribution function on $[a_i, b_i]$, then $F_i^{-1}(t)$ is the same as the parametric representation $I_i(t)$ in Eq.8. Thus, the Wasserstein distance coincides with the distance defined in Eq.10.

Now we are ready to construct a distance between fuzzy data. According to $\alpha$-cuts, the Wasserstein distance $d_{Wass}$ can be generalized to fuzzy numbers $\widetilde{A}_1$ and $\widetilde{A}_2$:

$$d(\widetilde{A}_1, \widetilde{A}_2) = \left\{ \int_0^1 d^2_{Wass}\big((\widetilde{A}_1)_\alpha, (\widetilde{A}_2)_\alpha\big)\, d\alpha \right\}^{\frac{1}{2}}. \tag{12}$$

We calculate this distance for triangular fuzzy numbers. Let $\widetilde{A}_i = (c_i, l_i, r_i)$, $i = 1, 2$ be triangular fuzzy numbers and $(\widetilde{A}_i)_\alpha = [l_i\alpha + (c_i - l_i) \;\; -r_i\alpha + (c_i + r_i)]$, the midpoint and the radius of $(\widetilde{A}_i)_\alpha$ are as follows:

$$m_{(\widetilde{A}_i)_\alpha} = c_i + \frac{1}{2}(1-\alpha)(r_i - l_i). \tag{13}$$

$$\delta_{(\widetilde{A}_i)_\alpha} = \frac{1}{2}(1-\alpha)(r_i + l_i). \tag{14}$$

Then we have:

$$\begin{aligned} d^2(\widetilde{A}_1, \widetilde{A}_2) &= \int_0^1 d^2_{Wass}(\,(\widetilde{A}_1)_\alpha, (\widetilde{A}_2)_\alpha)\, d\alpha \\ &= \int_0^1 \left\{ [m_{(\widetilde{A}_1)_\alpha} - m_{(\widetilde{A}_2)_\alpha}]^2 + \frac{1}{3}[\delta_{(\widetilde{A}_1)_\alpha} - \delta_{(\widetilde{A}_2)_\alpha}]^2 \right\} d\alpha \\ &= \int_0^1 \left\{ \left( (c_1 - c_2) + \frac{1}{2}(1-\alpha)\big[(r_1 - r_2) - (l_1 - l_2)\big] \right)^2 \right. \\ &\left. + \frac{1}{12}(1-\alpha)^2\big[(r_1 - r_2) + (l_1 - l_2)\big]^2 \right\} d\alpha \\ &= (c_1 - c_2)^2 + \frac{1}{9}\left[(l_1 - l_2)^2 + (r_1 - r_2)^2 - (l_1 - l_2)(r_1 - r_2)\right] \\ &- \frac{1}{2}(c_1 - c_2)\left[(l_1 - l_2) - (r_1 - r_2)\right]. \end{aligned} \tag{15}$$

We can use the distance (15) to define a distance between any two vectors of fuzzy numbers, by considering the sum of squared distances between individual elements. [See equation (20) ahead for more details.] In the next section, this distance is used for fuzzy clustering of fuzzy data.

## 5 Fuzzy clustering of fuzzy data with outliers

In this section we propose two approaches. In the first approach, based on our distance, we propose a fuzzy clustering model for fuzzy data, by modifying Keller's algorithm [22]. In the second approach, by defining a transformation, we reduce the problem of fuzzy clustering of fuzzy data to fuzzy clustering of crisp data. With the help of the second approach, any fuzzy clutering algorithms for crisp data can be used for fuzzy clustering of fuzzy data. For the sake of comparison with the first approach, we again use Keller's algorithm. Before describing the approaches, let us introduce some notations.

Let $U \equiv \{u_{ik} : i = 1, \ldots, c;\ k = 1, \ldots, n\}$ be the $c \times n$ membership matrix, where $c$ is the number of clusters, $n$ the number of data vectors and $u_{ik} \in [0, 1]$ the membership degree of the $k$-th object to the $i$-th cluster. We consider each data point, denoted as $\widetilde{x}_k$, and each cluster prototype, denoted as $\widetilde{v}_i$, to be a $p$-dimensional vector of triangular fuzzy data. This is in contrast to the Keller's approach where data elements and cluster prototypes are crisp.

To be more specific, let $\widetilde{x}_{kj}$ denote the $j$-th component of $\widetilde{x}_k$, the $k$-th data point. Then, $\widetilde{x}_{kj}$ can be represented as a 3-vector collecting its center, left spread and right spread. In symbols, we have

$$\widetilde{x}_{kj} := \left[c_{\widetilde{x}_{kj}}\ l_{\widetilde{x}_{kj}}\ r_{\widetilde{x}_{kj}}\right]^T \in \mathbb{R}^3, \tag{16}$$

$$\widetilde{x}_k := \left[\widetilde{x}_{k1}^T\ \widetilde{x}_{k2}^T\ \cdots\ \widetilde{x}_{kp}^T\right]^T \in \mathbb{R}^{3p}, \tag{17}$$

for $k = 1, \ldots, n$. In other words, we may view each data point, $\widetilde{x}_k$, either as a $p$-dimensional vector of fuzzy elements $\widetilde{x}_{kj}$ or as a $3p$-dimensional vector of real numbers. Both viewpoints are helpful and will be used interchangeably in what follows. A similar representation will be used for cluster prototypes, $\widetilde{v}_i$. That is,

$$\widetilde{v}_{ij} := \left[c_{\widetilde{v}_{ij}}\ l_{\widetilde{v}_{ij}}\ r_{\widetilde{v}_{ij}}\right]^T \in \mathbb{R}^3, \tag{18}$$

$$\widetilde{v}_i := \left[\widetilde{v}_{i1}^T\ \widetilde{v}_{i2}^T\ \cdots\ \widetilde{v}_{ip}^T\right]^T \in \mathbb{R}^{3p}, \tag{19}$$

for $i = 1, \ldots, c$.

As mentioned earlier, we consider the following (squared) distance between fuzzy vectors $\widetilde{x}_k$ and $\widetilde{v}_i$,

$$d^2(\widetilde{v}_i, \widetilde{x}_k) = \sum_{j=1}^{p} d^2(\widetilde{v}_{ij}, \widetilde{x}_{kj}), \tag{20}$$

where, $d^2(\widetilde{v}_{ij}, \widetilde{x}_{kj})$ is the (squared) distance (15) between fuzzy numbers $\widetilde{v}_{ij}$ and $\widetilde{x}_{kj}$.

### 5.1 Approach I

Following Keller, we minimize the objective function:

$$\mathcal{J}(U, \widetilde{V}; \widetilde{X}) = \sum_{i=1}^{c} \sum_{k=1}^{n} u_{ik}^m \cdot \frac{1}{\omega_k^q} \cdot d^2(\widetilde{v}_i, \widetilde{x}_k). \tag{21}$$

subject to the constraints

$$\sum_{k=1}^{n} \omega_k = \omega, \tag{22}$$

$$\sum_{i=1}^{c} u_{ik} = 1, \tag{23}$$

where, $m$ is the degree of fuzziness and $d^2(\widetilde{v}_i, \widetilde{x}_k)$ is as defined in (20).

The factor $\omega_k$ represents the weight of the $k$th datum and $\omega$ is a constant real valued parameter. According to Keller, the introduction of these weight factors helps in identifying outliers and reducing their effects. With constant parameter $q$, the influence of the outlier weight factors can be controlled. For this purpose, outliers are assigned a large weight $\omega_k$, so $\frac{1}{\omega_k^q}$ is small in this case.

The necessary conditions for minimizing the objective function are as follows:

$$c_{\widetilde{v}_{ij}} = \frac{\sum_{k=1}^{n} u_{ik}^m \cdot \frac{1}{\omega_k^q}\left[2c_{\widetilde{x}_{kj}} + \frac{1}{2}\big[(l_{\widetilde{v}_{ij}} - l_{\widetilde{x}_{kj}}) - (r_{\widetilde{v}_{ij}} - r_{\widetilde{x}_{kj}})\big]\right]}{2\sum_{k=1}^{n} u_{ik}^m \cdot \frac{1}{\omega_k^q}}. \tag{24}$$

$$l_{\widetilde{v}_{ij}} = \frac{\sum_{k=1}^{n} u_{ik}^m \cdot \frac{1}{\omega_k^q}\left[\frac{2}{9} l_{\widetilde{x}_{kj}} + \frac{1}{9}(r_{\widetilde{v}_{ij}} - r_{\widetilde{x}_{kj}}) + \frac{1}{2}(c_{\widetilde{v}_{ij}} - c_{\widetilde{x}_{kj}})\right]}{\frac{2}{9}\sum_{k=1}^{n} u_{ik}^m \cdot \frac{1}{\omega_k^q}}. \tag{25}$$

$$r_{\widetilde{v}_{ij}} = \frac{\sum_{k=1}^{n} u_{ik}^m \cdot \frac{1}{\omega_k^q}\left[\frac{2}{9} r_{\widetilde{x}_{kj}} + \frac{1}{9}(l_{\widetilde{v}_{ij}} - l_{\widetilde{x}_{kj}}) - \frac{1}{2}(c_{\widetilde{v}_{ij}} - c_{\widetilde{x}_{kj}})\right]}{\frac{2}{9}\sum_{k=1}^{n} u_{ik}^m \cdot \frac{1}{\omega_k^q}}. \tag{26}$$

$$\omega_k = \frac{\left(\sum_{i=1}^{c} u_{ik}^m \,.\, d^2(\widetilde{v}_i, \widetilde{x}_k)\right)^{\frac{1}{q+1}}}{\sum_{l=1}^{n}\left(\sum_{i=1}^{c} u_{il}^m \,.\, d^2(\widetilde{v}_i, \widetilde{x}_l)\right)^{\frac{1}{q+1}}} \cdot \omega. \tag{27}$$

$$u_{ik} = \frac{1}{\sum_{r=1}^{c}\left(\frac{d^2(\widetilde{v}_i, \widetilde{x}_k)}{d^2(\widetilde{v}_r, \widetilde{x}_k)}\right)^{\frac{1}{m-1}}}. \tag{28}$$

As it is observed, the membership degrees are left unchanged, while the cluster centers take into account the weights; points with high representativeness are more effective than outliers. On the basis of the necessary conditions, we can construct an iterative algorithm as follows:

---

**Algorithm:**

**Step 1.** Fix the degree of fuzziness ($m$), the number of clusters ($c$), $\omega$ and $q$. Choose an initial fuzzy c-partition $U^{(0)}$. Also, choose initial spreads and weights for each datum subject to Eq.(22). Set t=0.

**Step 2.** Calculate $\widetilde{V}^{(t)} = (c_{\widetilde{v}}^{(t)}, l_{\widetilde{v}}^{(t)}, r_{\widetilde{v}}^{(t)})$ using $U^{(t)}$, spreads, weights and Eqs.(24-26)

**Step 3.** Update $\omega_k^{(t)}$, $k = 1, \cdots, n$ using Eq.(27) and update $U^{(t)}$ by $U^{(t+1)}$ using $\widetilde{V}^{(t)} = (c_{\widetilde{v}}^{(t)}, l_{\widetilde{v}}^{(t)}, r_{\widetilde{v}}^{(t)})$ and Eq.(28)

**Step 4.** If $\|U^{(t+1)} - U^{(t)}\| < \varepsilon$ , where $\varepsilon$ is a non-negative small number fixed by the researcher, the algorithm has converged. Otherwise, set $t = t + 1$ and go to step 2.

---

### 5.2 Approach II

This approach is based on a different view of the distance (20). With some linear algebra, one can reduce this distance to the usual $3p$-dimensional Euclidean distance. For any $N$-vector, say $y = (y_1, \ldots, y_N) \in \mathbb{R}^N$, let $\|y\|_2 := \left(\sum_{i=1}^N y_i^2\right)^{1/2}$ denote its Euclidean norm.

Consider two triangular fuzzy numbers $\widetilde{A}_i = (c_i, l_i, r_i)$, $i = 1, 2$. Letting $c = c_1 - c_2$, $l = l_1 - l_2$, $r = r_1 - r_2$ and $z = [c,\ l,\ r]^T$, Eq. (15) can be rewritten as:

$$d^2(\widetilde{A}_1, \widetilde{A}_2) = c^2 + \frac{1}{9}l^2 + \frac{1}{9}r^2 - \frac{1}{9}lr - \frac{1}{2}cl + \frac{1}{2}cr \tag{29}$$

or equivalently as:

$$d^2(\widetilde{A}_1, \widetilde{A}_2) = [c,\ l,\ r] \begin{bmatrix} 1 & -\frac{1}{4} & \frac{1}{4} \\ -\frac{1}{4} & \frac{1}{9} & -\frac{1}{18} \\ \frac{1}{4} & -\frac{1}{18} & \frac{1}{9} \end{bmatrix} \begin{bmatrix} c \\ l \\ r \end{bmatrix} \tag{30}$$

Let us denote the matrix above as $Q$. The eigenvalues of Q are $\lambda_1 = \frac{7+\sqrt{43}}{12}$, $\lambda_2 = \frac{1}{18}$ and $\lambda_3 = \frac{7-\sqrt{43}}{12}$. Since Q is a real symmetric matrix, it is diagonalizable by orthogonal matrices. That is, there is an orthogonal $3 \times 3$ matrix $U$ (whose columns are orthonormal eigenvectors of $Q$) for which we have

$$Q = U \underbrace{\begin{bmatrix} \lambda_1 & 0 & 0 \\ 0 & \lambda_2 & 0 \\ 0 & 0 & \lambda_3 \end{bmatrix}}_{\Lambda} U^T. \tag{31}$$

Let $T$ be the (symmetric) square root of $Q$, i.e.,

$$T := Q^{1/2} = U\Lambda^{1/2}U^T. \tag{32}$$

Then, we may write

$$d(\widetilde{A}_1, \widetilde{A}_2) = \sqrt{z^T Q z} = \sqrt{z^T Q^{1/2} Q^{1/2} z} = \sqrt{z^T T^T T z} = \|Tz\|_2. \tag{33}$$

Now, recalling definitions (16) and (18) of $\widetilde{x}_{kj}$ and $\widetilde{v}_{ij}$, consider the following transformations

$$\widehat{x}_{kj} := T\,\widetilde{x}_{kj}, \quad \widehat{v}_{ij} := T\,\widetilde{v}_{ij} \tag{34}$$

where $\widetilde{x}_{kj}$ and $\widetilde{v}_{ij}$ are treated as 3-vectors. Furthermore, let us stack $\{\widehat{x}_{kj}\}_j$ and $\{\widehat{v}_{ij}\}_j$ into $3p$-dimensional vectors as usual, i.e.,

$$\widehat{x}_k := \begin{bmatrix} \widehat{x}_{k1}^T & \widehat{x}_{k2}^T & \cdots & \widehat{x}_{kp}^T \end{bmatrix}^T \in \mathbb{R}^{3p},$$
$$\widehat{v}_i := \begin{bmatrix} \widehat{v}_{i1}^T & \widehat{v}_{i2}^T & \cdots & \widehat{v}_{ip}^T \end{bmatrix}^T \in \mathbb{R}^{3p}.$$

Combining (20), (33) and the definitions of $\widehat{x}_k$ and $\widehat{v}_k$, we obtain

$$d(\widetilde{x}_k, \widetilde{v}_i) := \sqrt{\sum_{j=1}^{p} d^2(\widetilde{x}_{kj}, \widetilde{v}_{ij})} = \sqrt{\sum_{j=1}^{p} \|\widehat{x}_{kj} - \widehat{v}_{ij}\|_2^2} = \|\widehat{x}_k - \widehat{v}_i\|_2. \tag{35}$$

Equation (35) shows that the distance between fuzzy vectors $\widetilde{x}_k$ and $\widetilde{v}_i$ is the same as the Euclidean distance between the transformed vectors $\widehat{x}_k$ and $\widehat{v}_i$. In other words, we have reduced the problem of fuzzy clustering of fuzzy data to fuzzy clustering of crisp data. Thus, after applying transformations (34), any fuzzy clutering algorithm for crisp data can be used.

In particular, we can directly apply Keller's algorithm to $\{\widehat{x}_k\}$ by minimizing the objective function

$$\mathcal{J}(U, \widehat{V}; \widehat{X}) = \sum_{i=1}^{c} \sum_{k=1}^{n} u_{ik}^m \cdot \frac{1}{\omega_k^q}\ \|\widehat{x}_k - \widehat{v}_i\|_2^2, \tag{36}$$

under the same constraints (22) and (23) on $\{\omega_k\}$ and $\{u_{ik}\}$.

Necessary conditions for minimizing (36) are as follows:

$$\widehat{v}_i = \frac{\displaystyle\sum_{k=1}^{n} u_{ik}^m \cdot \frac{1}{\omega_k^q} \widehat{x}_k}{\displaystyle\sum_{k=1}^{n} u_{ik}^m \cdot \frac{1}{\omega_k^q}}. \tag{37}$$

$$\omega_k = \frac{\left(\displaystyle\sum_{i=1}^{c} u_{ik}^m \,.\, d^2(\widehat{v}_i, \widehat{x}_k)\right)^{\frac{1}{q+1}}}{\displaystyle\sum_{l=1}^{n} \left(\sum_{i=1}^{c} u_{il}^m \,.\, d^2(\widehat{v}_i, \widehat{x}_l)\right)^{\frac{1}{q+1}}} \cdot \omega. \tag{38}$$

$$u_{ik} = \frac{1}{\displaystyle\sum_{r=1}^{c} \left(\frac{d^2(\widehat{v}_i, \widehat{x}_k)}{d^2(\widehat{v}_r, \widehat{x}_k)}\right)^{\frac{1}{m-1}}}. \tag{39}$$

After iterations, equation (37) provides cluster prototypes in the transformed domain. To retrieve the fuzzy prototypes, one should apply the inverse transformation $T^{-1}$, i.e. $\widetilde{v}_{ij} = T^{-1}\widehat{v}_{ij}$, for $i = 1, \ldots, c$ and $j = 1, \ldots, p$.

## 6 Simulation experiments

In order to show how well our method works, two simulation experiments are conducted; one in an environment without outliers and the other one in presence of outliers. We almost obtained the same results with both approaches.

### 6.1 Clustering fuzzy data without outliers

We now discuss the results of a simulation study carried out in order to compare the performance of our model with existing models able to handle fuzzy data. These models are proposed by D'Urso and Giordani [11], by Yang et al. [36], by Hathaway et al. [18] and by Yang and Liu [38]. In order to compare the models, 2160 fuzzy data sets were randomly generated. After running several models for different values of $q$ and $\omega$, we chose $q = 1$, $\omega = 200$. The other parameters for clustering algorithm were set as follows: Number of objects ($n = 10, 50, 100$), number of variables ($k = 2, 8, 16$) and the weighting exponent ($m = 2, 3$). We constructed the data sets in such a way that $c = 2$ patterns can be found all over the simulation. To this purpose, the centers corresponding to the first $n/2$ objects were generated from the uniform distribution in $[0, 1]$, and those corresponding to the latter $n/2$ from the uniform distribution in $[0 + \theta, 1 + \theta]$. All the spreads were generated from the uniform distribution in $[0, 1]$ (case $\alpha$). On the other hand, in case $\beta$, all the centers were generated from the uniform distribution in $[0, 1]$, while the spreads corresponding to the first $n/2$ objects were generated from the uniform distribution in $[0, 1]$, and those corresponding to the latter $n/2$ from the uniform distribution in $[0 + \theta, 1 + \theta]$. $\theta$ was set to 1.5 and 0.75. In case of $\theta = 1.5$, the clusters are separated, whereas they are overlapped when $\theta$ is set to 0.75. Moreover, three sizes of centers with respect to the ones of the spreads were considered by defining a parameter h having the values 1/2,1,2. This parameter means that the size of the spreads is h times that of the centers.

In tables 1 and 2, the percentage of well-classified objects by the models are given. This is done by fixing one parameter at a time and averaging over the rest. So, the left columns of table 1 and 2 display the fixed parameters. Inasmuch as the cluster membership functions were known in advance, it is presumed that an object is assigned to a cluster correctly if the membership degree was the highest among all ($u = 0.5$). In addition, membership degrees higher than 0.75 and 0.9 are reported so that the strength of our model can be evaluated . It can be seen that our model works better in most of the conditions. As a case in point, when $\theta = 1.5$ and in conditions $n = 10, 50, 100$, $k = 2, 8, 16$, $\alpha$, $m = 2, 3$, our model had better performance for $u = 0.5, 0.75, 0.9$. When $\theta = 0.75$ and in conditions $n = 10, 50, 100$, $k = 2, 8, 16$, $m = 2, 3$, our model worked better, whereas in case $\beta$ the models proposed by Hathaway et al. and Yang and Liu, had the best performance. As reported in the tables 3,4, when $\theta = 1.5$, the average percentage of well-classified objects for our model is 97.39 ($u = 0.5$), 71.38 ($u = 0.75$) and 33.69 ($u = 0.9$). The model proposed by D'Urso and Giordini had the second highest performance after our model with 93.25 ($u = 0.5$), 60.81 ($u = 0.75$), 27.93 ($u = 0.9$). When $\theta$ passed from 1.5 to 0.75, the average performance of all models got worse. In this case, the average percentage of well-classified objects for our model is 91.12 ($u = 0.5$), 37.70 ($u = 0.75$) and 11.09 ($u = 0.9$) and for the model proposed by D'Urso and Giordini is 89.00 ($u = 0.5$), 32.56 ($u = 0.75$) and 7.35 ($u = 0.9$). As mentioned earlier, the simulation study showed that our model had much better results than the other existing models.

### 6.2 Clustering fuzzy data with Outliers

In order to evaluate how our model is able to detect the prototypes in case of possible presence of observations that can be seen as outliers, we added some outliers to cases $\alpha$ and $\beta$, mentioned above. After running several models for different values of $q$ and $\omega$, we chose $q = 2$, $\omega = 200$. The other parameters for clustering algorithm were set as follows: Number of objects ($n = 100, 200, 300$), where n/10 of them are outliers and the rest of the them are inliers, number of variables $k = (2, 8, 16)$ and the weighting exponent ($m = 2$). The modified cases $\alpha$ and $\beta$ are as follows:

- case $\alpha$: The centers corresponding to the first 1/2 of inliers were generated from the uniform distribution in $[0, 1]$, and those corresponding to the rest of the inliers from the uniform distribution in $[1.5, 2.5]$. The number of outliers is n/10. The centers of outliers were generated from Normal distribution with mean=-2 and variance=2. The left and the right spreads were generated from the uniform distribution in $[0, 1]$
- case $\beta$: The left and the right spreads corresponding to the first 1/2 of inliers were generated from the uniform distribution in $[0, 1]$, and those corresponding to the rest of the inliers from the uniform distribution in $[1.5, 2.5]$. The number of outliers is n/10. The left and the right spreads of outliers were generated from Normal distribution with mean=5 and variance=2. All the centers were generated from the uniform distribution in $[0, 1]$.

The mean square errors (MSE) between prototypes obtained by performing our clustering model and the ideal prototypes are shown in tables 5 and 6. From these tables, it can be observed that MSE of the centers are more than MSE of left spreads and right spreads in case $\alpha$, while MSE of spreads are higher than those of centers in case $\beta$. In both cases, small weighting factors $\omega_k$ (large values for $\frac{1}{\omega_k^q}$) are assigned to data points fitting well to one of the clusters whereas large $\omega_k$ (small values for $\frac{1}{\omega_k^q}$) are assigned to outliers. Thus, outliers can be easily identified by their large weighting factors.

## 7 Conclusion and future works

This paper presented a fuzzy clustering model for fuzzy data based on a new distance. We have modified Keller's approach so that our model can be used in noisy environments. The weighting factors reduce the influence of outliers and enable us to identify them. Necessary conditions for the objective function to receive an optimum have been derived to calculate a partition of data. Also, in another approach, we transformed our distance to the Euclidean distance and reduced the problem of fuzzy clustering of fuzzy data to fuzzy clustering of crisp data. Finally, two simulation experiments were considered; one for comparing the performance of our model with those of other existing clustering models for fuzzy data and one for testing how well our model behaves in noisy environments.

Our model can be applied in settings where the presence of outliers can drastically affect the results. An example is the process control problem in which the presence of outliers usually represents that the process has been out of control.

Another problem that can be explored is to study in depth fuzzy clustering for interactive fuzzy data and determining the optimal weighting exponent (m).

Table 1: Percentages of well-classified objects with membership higher than $u = 0.5$, $u = 0.75$, $u = 0.9$ ($\theta = 1.5$)

| | Our model | | | D'Urso and Giordani (2007) | | | Hathaway et. al. (1996) Yang and Liu (1999) | | | Yang et. al. (2004) | | |
|---|---|---|---|---|---|---|---|---|---|---|---|---|
| | $u = 0.5$ | $u = 0.75$ | $u = 0.9$ | $u = 0.5$ | $u = 0.75$ | $u = 0.9$ | $u = 0.5$ | $u = 0.75$ | $u = 0.9$ | $u = 0.5$ | $u = 0.75$ | $u = 0.9$ |
| $n = 10$ | 97.53 | 74.94 | 38.14 | 94.11 | 66.22 | 31.78 | 94.78 | 56.61 | 19.11 | 84.83 | 54.94 | 24.39 |
| $n = 50$ | 97.86 | 70.14 | 32.03 | 94.18 | 58.47 | 26.14 | 94.34 | 48.36 | 15.32 | 85.97 | 47.81 | 19.74 |
| $n = 100$ | 97.68 | 69.12 | 31.16 | 91.45 | 57.74 | 25.87 | 92.88 | 47.21 | 14.82 | 85.57 | 46.27 | 18.90 |
| $k = 2$ | 93.33 | 73.17 | 38.02 | 84.36 | 64.74 | 31.08 | 85.05 | 53.29 | 18.16 | 82.18 | 55.59 | 24.33 |
| $k = 8$ | 98.95 | 70.86 | 31.89 | 96.11 | 59.95 | 27.15 | 98.26 | 49.17 | 15.67 | 86.50 | 47.80 | 19.59 |
| $k = 16$ | 99.68 | 70.21 | 31.11 | 99.27 | 58.73 | 25.56 | 98.69 | 49.71 | 15.42 | 87.69 | 45.64 | 19.11 |
| $\alpha$ | 100.00 | 98.83 | 52.63 | 100.00 | 97.30 | 51.32 | 93.36 | 50.91 | 16.47 | 100.00 | 87.29 | 39.82 |
| $\beta$ | 94.78 | 43.69 | 15.33 | 86.49 | 24.32 | 4.54 | 94.64 | 50.54 | 16.37 | 70.91 | 12.06 | 2.20 |
| $m = 2$ | 97.13 | 81.64 | 62.27 | 93.12 | 71.75 | 53.56 | 94.34 | 69.79 | 32.62 | 85.50 | 60.51 | 40.90 |
| $m = 3$ | 97.49 | 60.48 | 4.43 | 93.37 | 49.87 | 2.30 | 93.66 | 31.66 | 0.21 | 85.41 | 38.84 | 1.13 |
| $h = 1/2$ | 91.94 | 55.20 | 29.02 | 91.96 | 51.79 | 25.44 | 91.96 | 51.31 | 21.95 | 76.28 | 54.23 | 27.59 |
| $h = 1$ | 99.72 | 70.65 | 29.88 | 100.00 | 74.77 | 28.33 | 99.94 | 50.65 | 5.29 | 82.29 | 52.24 | 25.01 |
| $h = 2$ | 100.00 | 88.96 | 42.06 | 87.78 | 55.87 | 30.01 | 90.10 | 50.21 | 22.01 | 97.79 | 42.55 | 10.43 |

Table 2: Percentages of well-classified objects with with membership higher than $u = 0.5$, $u = 0.75$, $u = 0.9$ ($\theta = 0.75$)

| | Our model | | | D'Urso and Giordani (2007) | | | Hathaway et. al. (1996) Yang and Liu (1999) | | | Yang et. al. (2004) | | |
|---|---|---|---|---|---|---|---|---|---|---|---|---|
| | $u = 0.5$ | $u = 0.75$ | $u = 0.9$ | $u = 0.5$ | $u = 0.75$ | $u = 0.9$ | $u = 0.5$ | $u = 0.75$ | $u = 0.9$ | $u = 0.5$ | $u = 0.75$ | $u = 0.9$ |
| $n = 10$ | 89.86 | 47.72 | 17.47 | 89.78 | 39.89 | 12.39 | 88.72 | 32.28 | 6.06 | 81.89 | 37.39 | 10.33 |
| $n = 50$ | 91.34 | 35.28 | 8.23 | 89.43 | 29.17 | 5.07 | 87.87 | 20.81 | 1.73 | 79.39 | 28.37 | 4.71 |
| $n = 100$ | 92.26 | 32.08 | 7.19 | 87.78 | 28.86 | 4.58 | 86.27 | 19.91 | 1.61 | 80.20 | 27.63 | 4.76 |
| $k = 2$ | 82.82 | 44.91 | 18.49 | 81.40 | 41.03 | 13.75 | 78.98 | 32.25 | 7.67 | 74.43 | 39.50 | 11.97 |
| $k = 8$ | 93.25 | 35.11 | 8.96 | 88.95 | 29.18 | 4.99 | 89.11 | 21.31 | 1.12 | 82.85 | 27.67 | 4.41 |
| $k = 16$ | 97.17 | 32.49 | 6.53 | 96.64 | 27.71 | 3.29 | 94.78 | 19.43 | 0.61 | 84.20 | 26.22 | 3.44 |
| $\alpha$ | 98.52 | 60.97 | 18.29 | 98.97 | 55.20 | 12.57 | 86.33 | 24.77 | 2.93 | 98.99 | 54.97 | 11.39 |
| $\beta$ | 83.32 | 14.01 | 3.15 | 79.02 | 10.07 | 2.12 | 88.91 | 23.89 | 3.34 | 61.99 | 7.28 | 1.81 |
| $m = 2$ | 90.58 | 59.62 | 20.41 | 89.14 | 55.50 | 14.44 | 87.89 | 43.68 | 6.18 | 80.64 | 50.71 | 12.95 |
| $m = 3$ | 91.94 | 16.16 | 1.95 | 88.86 | 9.77 | 0.25 | 87.35 | 4.99 | 0.09 | 80.34 | 11.54 | 0.25 |
| $h = 1/2$ | 83.88 | 38.21 | 14.67 | 84.75 | 29.75 | 4.04 | 84.75 | 29.75 | 4.04 | 75.21 | 38.71 | 11.62 |
| $h = 1$ | 91.75 | 35.84 | 11.44 | 97.62 | 31.66 | 5.29 | 96.87 | 14.71 | 1.22 | 75.90 | 31.06 | 6.22 |
| $h = 2$ | 97.94 | 38.60 | 7.74 | 84.63 | 36.51 | 12.71 | 81.24 | 28.53 | 4.13 | 90.37 | 23.61 | 1.97 |

Table 3: Average percentage of well-classified objects ($\theta = 1.5$)

| | $u = 0.5$ | $u = 0.75$ | $u = 0.9$ |
|---|---|---|---|
| Our model | 97.39 | 71.38 | 33.69 |
| D'Urso and Giordani | 93.25 | 60.81 | 27.93 |
| Hathaway et.al. | 94.00 | 50.72 | 16.42 |
| Yang et.al. | 85.46 | 49.67 | 21.01 |

Table 4: Average percentage of well-classified objects ($\theta = 0.75$)

| | $u = 0.5$ | $u = 0.75$ | $u = 0.9$ |
|---|---|---|---|
| Our model | 91.12 | 37.70 | 11.09 |
| D'Urso and Giordani | 89.00 | 32.56 | 7.35 |
| Hathaway et.al. | 87.62 | 24.33 | 3.13 |
| Yang et.al. | 80.49 | 31.13 | 6.60 |

Table 5: Mean Square Error (MSE) for cluster prototypes - case($\alpha$)(with outliers)

| | Centers (MSE) | Left Spreads (MSE) | Right Spreads (MSE) |
|---|---|---|---|
| $n = 100$ | 0.0035 | 0.0015 | 0.0016 |
| $n = 200$ | 0.0025 | 0.0007 | 0.0007 |
| $n = 300$ | 0.0022 | 0.0005 | 0.0005 |
| $k = 2$ | 0.0025 | 0.0014 | 0.0013 |
| $k = 8$ | 0.0024 | 0.0007 | 0.0007 |
| $k = 16$ | 0.0034 | 0.0006 | 0.0006 |

Table 6: Mean Square Error (MSE) for cluster prototypes - case($\beta$)(with outliers)

| | Centers (MSE) | Left Spreads (MSE) | Right Spreads (MSE) |
|---|---|---|---|
| $n = 100$ | 0.0031 | 0.1365 | 0.1363 |
| $n = 200$ | 0.0014 | 0.1447 | 0.1441 |
| $n = 300$ | 0.0010 | 0.1481 | 0.1482 |
| $k = 2$ | 0.0033 | 0.0140 | 0.0140 |
| $k = 8$ | 0.0010 | 0.1232 | 0.1227 |
| $k = 16$ | 0.0010 | 0.2937 | 0.2932 |

## References

[1] Abonyi, J., Roubos, J.A., Szeifert, F., (2003). *Data-driven generation of compact, accurate, and linguistically sound fuzzy classifiers based on a decision-tree initialization.*Internat. J. Approx. Reason. 32, 1?21.

[2] Auephanwiriyakul, S., Keller, J.M., (2002). *Analysis and efficient implementation of a linguistic fuzzy c-means.* IEEE Trans. Fuzzy Systems 10 (5), 563?582.

[3] Berkhin, P. (2002). Survey of clustering data mining techniques. Accrue Software Inc. http://www.accrue.com/products/researchpapers. html.

[4] Bertoluzza, C., Corral, N., Salas, A. (1995).*On a new class of distance between fuzzy numbers*, Mathware and Soft Computing, 2, 71?84.

[5] Bezdek, J.C., (1981). *Pattern Recognition with Fuzzy Objective Function Algorithms.* Plenum Press, NewYork.

[6] Bezdek, J.C. (1974a). *Numerical taxonomy with fuzzy sets?, Journal of Mathematical Biology*, 1, 57?71.

[7] Bezdek, J.C. (1974b). *Cluster validity with fuzzy sets?, Journal of Cybernetics*, 9, 58?72.

[8] Bloch, I. (1999) *On fuzzy distances and their use in image processing under imprecision*, Pattern Recognition, 32, 1873?1895.

[9] De Oliveira, JV and Pedrycz, W. (2007). Advances in Fuzzy Clustering and its Applications. San Francisco: Wiley.

[10] Diamond, P., Kloeden, P., (1994). *Metrics Spaces of Fuzzy Sets. Theory and Applications.* World Scientific, Singapore.

[11] D'Urso,P., Giordani,P. (2006a).*A weighted fuzzy c-means clustering model for fuzzy data*, Computational Statistics Data Analysis, 50 (6), 1496?1523.

[12] D'Urso, P., Giordani, P. (2006b). *A robust fuzzy k-means clustering model for interval valued data*, Computational Statistics, 21, 251?269.

[13] El-Sonbaty,Y., Ismail, M.A. (1998). *Fuzzy Clustering for Symbolic Data*, IEEE Transactions on Fuzzy Systems, 6 (2),195?204.

[14] Gibbs, A.L., Su, F.E., (2002). *On choosing and bounding probability metrics*, Internat. Statist. Rev. 70, 419.

[15] Gowda, K.C., Diday, E., (1991). *Symbolic clustering using a new dissimilarity measure.* Pattern Recognition 24 (6), 567?578.

[16] Gowda, K.C., Diday, E., (1992). *Symbolic clustering using a new similarity measure.* IEEE Trans. Systems Man Cybern. 22, 368?378.

[17] Grzegorzewski, P. (2004). *Distance between intuitionistic fuzzy sets and/or interval-valued fuzzy sets based on the Hausdorff metric*, Fuzzy Sets and Systems, 148, 319?328.

[18] Hathaway, R.J., Bezdek, J.C., Pedrycz, W. (1996) *A parametric model for fusing heterogeneous fuzzy data*, IEEE Transactions on Fuzzy Systems, 4 (3), 1277?1282.

[19] Hung, W.-L., Yang, M.-S., (2004). *Similarity measures of intuitionistic fuzzy sets based on the Hausdorff metric*. Pattern Recognition Lett. 1603?1611.

[20] Hung, W.-L., Yang, M.-S., (2005). *Fuzzy clustering on LR-type fuzzy numbers with an application in Taiwanese tea evaluation*. Fuzzy Sets and Systems 150 (3), 561?577.

[21] Irpino, A., Verde, R. (2008).*Dynamic clustering for interval data using a Wasserstein-based distance*,Pattern Recognition Letters, 29, 1648-1658.

[22] Keller,A. (2000). *Fuzzy clustering with outliers* In Proc. of the 19th Int. Conf. of the North American Fuzzy Information Processing Society, NAFIPS00 (ed. Whalen T), pp. 143?147.

[23] Kim, D.S., Kim, Y.K., (2004). *Some properties of a new metric on the space of fuzzy numbers*. Fuzzy Sets and Systems 145, 395?410.

[24] Kown, S.H. (1998) *Cluster validity index for fuzzy clustering*IEEE electronic letters ,Vol. 34 No. 22.

[25] Li,Y.,Olson,D.L.,Qin,Z.(2007)*Similarity measures between intuitionistic fuzzy (vague) sets: a comparative analysis*, Pattern Recognition Letters, 28 (2), 278?285.

[26] Näther, W. (2000). *On random fuzzy variables of second order and their application to linear statistical inference with fuzzy data*, Metrika, 51, 201?221.

[27] Pal, N. R. and Bezdek, J.C., (1995). *On cluster validity for fuzzy c-means model*, IEEE Trans. Fuzzy Systems 1 , 370-379.

[28] Pappis, C.P., Karacapilidis, N.I., (1993).*A comparative assessment of measures of similarity of fuzzy values*. Fuzzy Sets and Systems 56, 171?174.

[29] Pedrycz, W., Bezdek, J.C., Hathaway, R.J., Rogers, G.W., (1998). *Two nonparametric models for fusing heterogeneous fuzzy data*. IEEE Trans. Fuzzy Systems 6 (3), 411?425.

[30] Rosenfeld, A., (1979). Fuzzy digital topology. Inform. Control 40, 76?87.

[31] Sato, M., Sato,Y., (1995). *Fuzzy clustering model for fuzzy data*. Proc. IEEE 2123?2128. bibitemszmidySzmidt, E., Kacprzyk, J., (2000). *Distances between intuitionistic fuzzy sets*. Fuzzy Sets and Systems 114, 505?518.

[32] Takata, O., Miyamoto, S., Umayahara, K., (1998). *Clustering of data with uncertainties using Hausdorff distance*. Proceedings of the 2nd IEEE International Conference on Intelligence Processing Systems, August 4?7, 1998, Gold Coast, Australia, pp. 67?71.

[33] Tran, L., Duckstein, L., (2002). *Comparison of fuzzy numbers using a fuzzy distance measure*, Fuzzy sets and systems, 130, 331-341

[34] Xie, X.L., Beni,G., (1991). *A validity measure for fuzzy clustering*, IEEE Trans. Pattern Anal. Machine Intelligence 13 841?847.

[35] Yang, M.S., Ko, C.H., (1996). *On a class of fuzzy c-numbers clustering procedures for fuzzy data*. Fuzzy Sets and Systems 84, 49?60.

[36] Yang, M.S., Hwang, P.Y., Chen, D.H. (2004). *Fuzzy clustering algorithms for mixed feature variables*, Fuzzy Sets and Systems, 141, 301?317.

[37] Yang,M.S.,Hwang,P.Y.,Chen,D.H.(2005). *On a similarity measure between LR-type fuzzy numbers and its application to database acquisition*, International Journal of Intelligent Systems, 20, 1001?1016.

[38] Yang, M.S., Liu, H.H. (1999). *Fuzzy clustering procedures for conical fuzzy vector data*, Fuzzy Sets and Systems, 106, 189?200.

[39] Yong,D.,Wenkang,S.,Feng,D.,Qi,L.(2004). *A new similarity measure of generalized fuzzy numbers and its application to pattern recognition*, Pattern Recognition Letters, 25, 875?883.

[40] Zhang, C., Fu, H. (2006). *Similarity measures on three kinds of fuzzy sets*, Pattern Recognition Letters, 27 (12), 1307?1317.

[41] Zhang, Y. , Wang, W., Zhang, X., Li, Y. (2008). A cluster valididty index for fuzzy clustering, Information Science 178,1205-1218.

[42] Zimmermann, H.J., (2001). *Fuzzy Set Theory and its Applications*. Kluwer Academic Press, Dordrecht.

[43] Zwich, R., Carlstein, E., Budescu, D.V., (1987). *Measures of similarity among fuzzy concepts: a comparative analysis*. Internat. Approx. Reason. 1, 221?242.